\documentclass[10pt,twocolumn,letterpaper]{article}

\usepackage[pagenumbers]{cvpr}      
\definecolor{cvprblue}{rgb}{0.21,0.49,0.74}
\usepackage[pagebackref,breaklinks,colorlinks,allcolors=cvprblue]{hyperref}
\usepackage{xcolor,colortbl}         
\usepackage{wrapfig}     
\usepackage{adjustbox}
\usepackage{amsfonts}
\usepackage{graphicx}

\def\paperID{15238} 
\def\confName{CVPR}
\def\confYear{2026}

\title{SwiftNDC: Fast Neural Depth Correction for High-Fidelity 3D Reconstruction}

\author{Kang Han$^{1}$ \quad Wei Xiang$^{1}$\thanks{Corresponding author.} \quad Lu Yu$^{1}$ \quad Mathew Wyatt$^{2}$ \quad Gaowen Liu$^{3}$ \quad Ramana Rao Kompella$^{3}$ \\
{\normalsize $^1$La Trobe University, Melbourne, VIC 3086, Australia}\\
{\normalsize $^2$Australian Institute of Marine Science, Perth, WA 6009, Australia}\\
{\normalsize $^3$Cisco Research, San Jose, CA, USA}\\
{\tt\small \{k.han, w.xiang, l.yu\}@latrobe.edu.au,}
{\tt\small m.wyatt@aims.gov.au,}
{\tt\small \{gaoliu, rkompell\}@cisco.com}
}

\newcommand{\SwiftNDC}{SwiftNDC\xspace}

\definecolor{yellow}{rgb}{1, 1, 0.7}
\definecolor{orange}{rgb}{1, 0.85, 0.7}
\definecolor{red}{rgb}{1, 0.7, 0.7}
\definecolor{normalred}{rgb}{1, 0, 0}

\begin{document}
\maketitle
\begin{abstract}
Depth-guided 3D reconstruction has gained popularity as a fast alternative to optimization-heavy approaches, yet existing methods still suffer from scale drift, multi-view inconsistencies, and the need for substantial refinement to achieve high-fidelity geometry. Here, we propose SwiftNDC, a fast and general framework built around a Neural Depth Correction field that produces cross-view consistent depth maps. From these refined depths, we generate a dense point cloud through back-projection and robust reprojection-error filtering, obtaining a clean and uniformly distributed geometric initialization for downstream reconstruction. This reliable dense geometry substantially accelerates 3D Gaussian Splatting (3DGS) for mesh reconstruction, enabling high-quality surfaces with significantly fewer optimization iterations. For novel-view synthesis, SwiftNDC can also improve 3DGS rendering quality, highlighting the benefits of strong geometric initialization. We conduct a comprehensive study across five datasets, including two for mesh reconstruction, as well as three for novel-view synthesis. SwiftNDC consistently reduces running time for accurate mesh reconstruction and boosts rendering fidelity for view synthesis, demonstrating the effectiveness of combining neural depth refinement with robust geometric initialization for high-fidelity and efficient 3D reconstruction.
\end{abstract}    
\section{Introduction}
\label{sec_introduction}

High-quality 3D reconstruction from multi-view images underpins a wide range of applications, including real-time rendering, robotics, simulation, content creation, and digital preservation \cite{farshian2023deep}. While recent radiance field methods such as Neural Radiance Fields (NeRF) \cite{mildenhall2020nerf,oechsle2021unisurf, li2023neuralangelo} and 3DGS \cite{kerbl20233d, guedon2024sugar, chen2024pgsr} have demonstrated impressive geometric and photometric fidelity, they typically require extensive optimization iterations per scene to converge to accurate geometry. This high computational cost poses a significant barrier for practical 3D reconstruction pipelines that must operate at scale.

In contrast, feed-forward depth estimation networks~\cite{ding2022transmvsnet, caomvsformer,caomvsformerpp, wang2025vggt,chen2025video} predict depth maps in seconds, offering an appealing route to fast scene reconstruction. However, even state-of-the-art depth predictors exhibit residual scale drift, local bias, and cross-view inconsistencies \cite{jensen2014large}. These small errors accumulate during back-projection and fusion, leading to degraded surface accuracy and unreliable geometry for downstream tasks such as mesh extraction or radiance field optimization. Despite extensive progress in depth prediction and refinement \cite{wang2025vggt,chen2025video,yan2023desnet, li2018megadepth, izquierdo2023sfm}, obtaining dense and multi-view consistent geometry at low computational cost remains an unsolved challenge.

We bridge this gap with \SwiftNDC, a unified framework that combines neural depth refinement with robust geometric filtering to produce high-quality dense geometry suitable for both mesh reconstruction and radiance-based representation learning. At the core of our approach is a Neural Depth Correction field that refines per-pixel depths from feed-forward networks using a lightweight and scene-specific model, producing depth maps that are geometrically aligned and consistent across views. These refined depths are back-projected into a dense point cloud, which we further clean through a multi-view reprojection-error filtering process. The result is a reliable, uniformly distributed dense geometry that can serve as a strong initialization for downstream reconstruction.

\begin{figure*}[t]
  \centering
  \includegraphics[width=\textwidth]{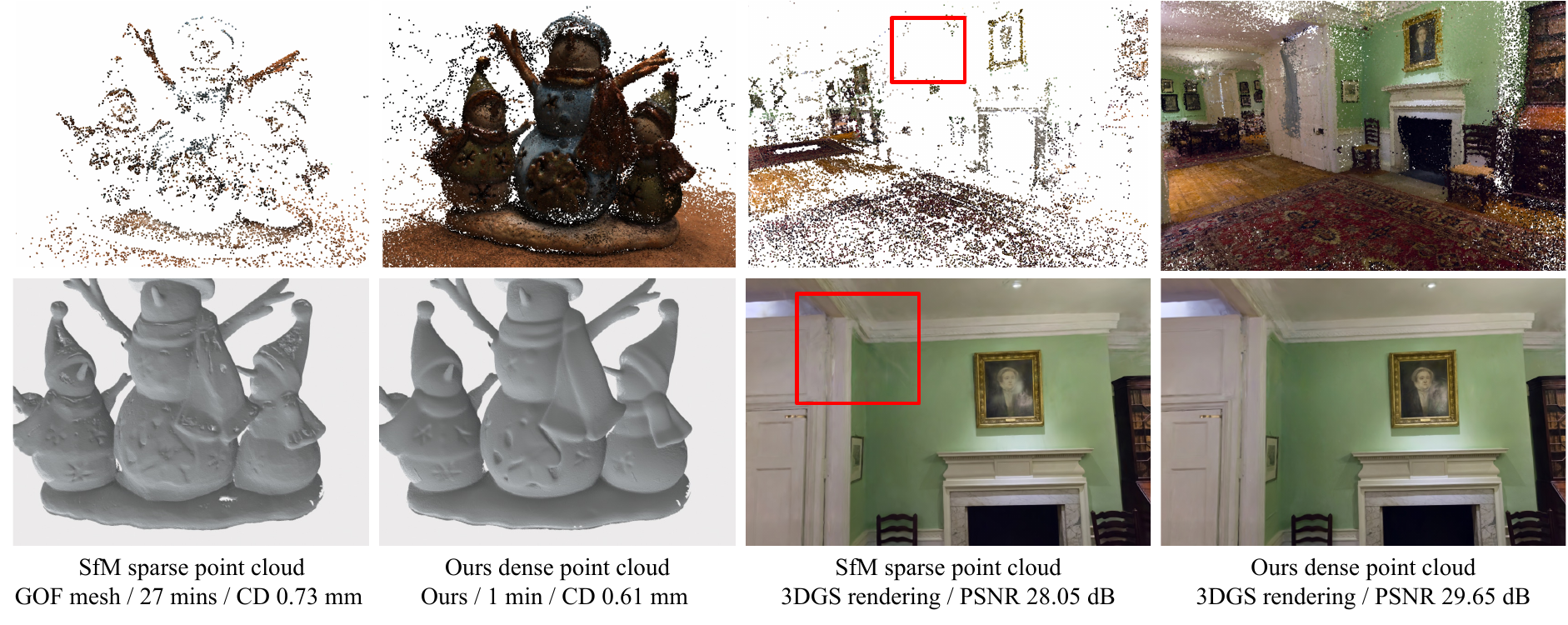}
  \caption{Comparison of mesh reconstruction and novel view synthesis. Standard 3DGS based methods (Gaussian Opacity Field (GOF)~\cite{yu2024gaussian}) initialized from a sparse SfM point cloud requires extensive optimization to achieve high-fidelity results. \SwiftNDC produces reliable dense geometry that accelerates mesh reconstruction (quality measured in Chamfer Distance (CD)$\downarrow$) and improves rendering quality for novel view synthesis.}
  \label{fig:mesh_intro}
\end{figure*}

This dense and accurate geometric initialization has two key benefits.
First, for surface mesh reconstruction, initializing 3DGS with our filtered dense geometry drastically reduces the required optimization iterations, achieving high-fidelity meshes in as few as 1k iterations on DTU object-level surface reconstruction. Second, for novel view synthesis, our initialization improves 3DGS rendering fidelity especially in weakly observed regions, highlighting the importance of accurate and well-distributed geometry for radiance-field learning (as shown in examples in Fig.\ref{fig:mesh_intro}). These complementary advantages allow \SwiftNDC to bridge the gap between fast depth-based pipelines and optimization-heavy radiance-field methods.

We conduct a comprehensive evaluation across five datasets, covering both surface reconstruction and novel view synthesis. \SwiftNDC consistently reduces reconstruction time for mesh reconstruction while improving rendering quality for novel view synthesis. In summary, our contributions are threefold:

\begin{itemize}
\item We introduce \SwiftNDC, a framework built upon a Neural Depth Correction field that yields geometrically accurate and cross-view consistent depth maps.
\item We propose a reliable dense geometry initialization method, combining depth back-projection with reprojection-error filtering to produce high-quality geometry for downstream reconstruction.
\item We present a comprehensive five-dataset evaluation, showing that \SwiftNDC reduces reconstruction time for mesh reconstruction and improves rendering quality for novel view synthesis.
\end{itemize}

\section{Related work}
\subsection{Per-scene optimization pipelines}
Neural implicit surface methods represent geometry as continuous fields, typically signed distance functions (SDFs) or density-radiance volumes encoded by multilayer perceptrons (MLPs) and optimized on a per-scene basis. Building on the idea of learning a SDF by an MLP in DeepSDF \cite{park2019deepsdf}, NeRF \cite{mildenhall2020nerf} introduced differentiable volumetric rendering of a density-radiance field, showing that photo-consistent geometry could be recovered without explicit 3D supervision. Subsequent methods such as NeuS \cite{wang2021neus}, VolSDF \cite{yariv2021volume}, UniSurf \cite{oechsle2021unisurf} and Neuralangelo \cite{li2023neuralangelo} refined this radiance geometry marriage by explicitly enforcing surface consistency while retaining view-dependent color. Although these models achieve sub-millimeter accuracy, they typically require hours of per-scene optimization because of intensive ray sampling and repeated network evaluations.

3DGS replaces volumetric representations with thousands of anisotropic Gaussians optimized via differentiable rasterization, dramatically reducing training time and accelerating rendering \cite{kerbl20233d}. Because the original 3DGS formulation optimizes only for photometric consistency, it lacks an explicit surface constraint, so extracting meshes from the raw Gaussian cloud yields noisy, incomplete geometry. To improve surface modeling, a line of variants has emerged, including SuGaR~\cite{guedon2024sugar}, 2DGS~\cite{huang20242d}, GOF~\cite{yu2024gaussian}, and NSDF-Splat \cite{zhangneural}. Although faster than neural implicit pipelines, these approaches still require extensive scene specific optimization, partially due to very sparse SfM initilization and iterative densification. EDGS \cite{kotovenko2025edgs} attempts to mitigate this cost through a dense Gaussian initialization obtained from triangulated 2D correspondences, but its design is tailored exclusively for novel view synthesis. In contrast, our approach produces accurate, multi-view-consistent dense geometry from corrected depth maps, enabling both high-fidelity surface mesh reconstruction and improved view synthesis.

\begin{figure*}[th]
  \centering
  \includegraphics[width=\textwidth]{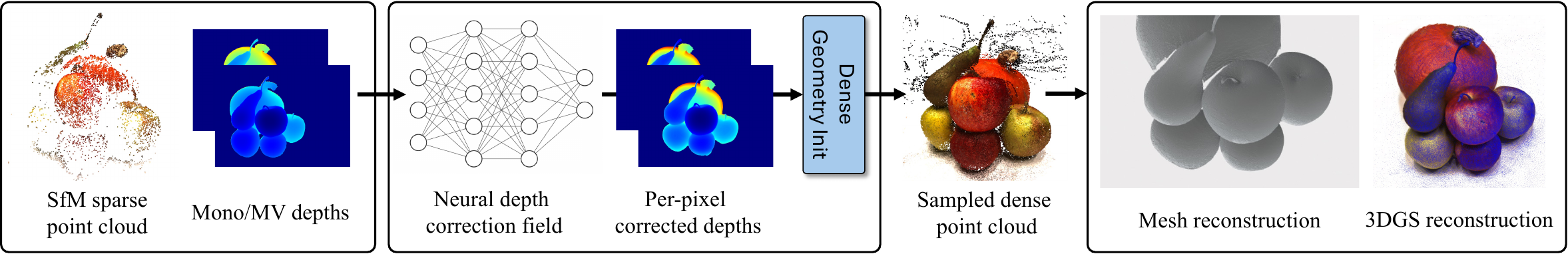}
  \caption{\SwiftNDC pipeline. The method starts with both monocular and multi-view depth maps. A neural depth correction field, supervised by sparse SfM points, refines these depths at pixel level. The corrected depths are then transformed into a dense and reliable sampled point cloud, which serves as a strong geometric initialization for downstream mesh reconstruction and 3DGS-based view synthesis.}
  \label{fig:overview}
\end{figure*}

\subsection{Feed-forward depth estimation}
Classical PatchMatch-style stereo algorithms, including COLMAP's MVS module \cite{schonberger2016structure}, propagate depth hypotheses across images but need several minutes per scene and struggle on texture-poor regions \cite{schonberger2016pixelwise}. Learning-based networks such as MVSNet \cite{yao2018mvsnet}, R-MVSNet \cite{yao2019recurrent}, Cas-MVSNet \cite{gu2020cascade}, TransMVSNet \cite{ding2022transmvsnet}, and the more recent MVSFormers \cite{caomvsformer,caomvsformerpp} replace hand-crafted costs with deep cost volumes and predict depth in seconds. However, small pixel-wise scale or bias errors still persist in MVS depth maps. After TSDF fusion \cite{curless1996volumetric} these residuals manifest as wavy surfaces, holes, and over-smoothed details \cite{wolf2024gs2mesh}. Because conventional feed-forward MVS optimizes each reference image independently, it is forced to subsample to about 3–7 best-overlapping views to keep cost volume memory affordable, sacrificing global geometric consistency and ultimately mesh fidelity.

Visual Geometry Grounded Transformer (VGGT) \cite{wang2025vggt} changes this landscape by ingesting all available views (up to hundreds) in a single forward pass and jointly predicting depths, cameras and point maps, thus sidestepping the fixed $K$ view bottleneck of current MVS networks. While VGGT's global multi-view reasoning yields depth maps that are markedly more 3D consistent, we observe persistent local mis-alignments and small metric biases that translate into poor, fragmented meshes when fused directly (see Fig.~\ref{fig:depth_pcd}). These residual errors motivate the pixel-level depth-correction strategy proposed in this work to refine VGGT outputs into high-fidelity dense geometry.

\subsection{Depth correction and test-time refinement}
Early depth alignment methods such as MegaDepth \cite{li2018megadepth} and AdaBins \cite{bhat2021adabins} applied a single scale factor or bias per image using sparse SfM depths, correcting gross metric error but not spatial variations. Recent work has moved toward pixel-level refinement. SfM-TTR \cite{izquierdo2023sfm} fine-tunes an entire depth network at test time from COLMAP points, while Fink et al. \cite{fink2024refinement} learn a residual MLP for the first image but then perform an additional differentiable rendering stage that adds extensive computation. Sparse-to-dense completion systems like RadarCam-Depth \cite{li2024radarcam} and depth from focus-based HybridDepth \cite{ganj2025hybriddepth} predict dense corrections with heavy CNN or Transformer backbones and biased on specific training datasets. All of these methods either operate on single images, require large network updates, or involve complex multi-stage optimization.

\SwiftNDC departs from prior work by learning a lightweight neural depth correction field that refines per-pixel depths using monocular and multi-view cues, supervised only by sparse SfM points. Crucially, the refined depths are converted into a dense point cloud and filtered via multi-view reprojection error, yielding reliable geometry for downstream reconstruction. This initialization enables high-fidelity meshes with fewer 3DGS iterations and improves novel view synthesis quality.

\section{Method}\label{sec:method}

\SwiftNDC efficiently converts an image set into multi-view-consistent depth maps and a reliable dense point cloud, completing this process in under one minute for scenes with fewer than 60 views. Fig.~\ref{fig:overview} depicts the proposed pipeline. Starting from COLMAP~\cite{schonberger2016structure} camera poses and sparse points, we obtain two dense depth estimates per view: a multi-view depth from VGGT~\cite{wang2025vggt}, which provides strong global consistency, and a monocular depth from Video Depth Anything (VDA)~\cite{chen2025video}, which preserves fine detail but is only up to scale. Both maps are coarsely aligned to the sparse points through a per-view affine fit. At the core of our approach is a pixel-level neural depth correction field trained with sparse SfM supervision, which removes the remaining local depth biases in under one second per image. The corrected depth maps are then back-projected to form a dense point cloud and refined via reprojection-error filtering to provide reliable geometry for downstream mesh reconstruction and view synthesis.

\begin{figure*}[t]
  \centering
  \includegraphics[width=\textwidth]{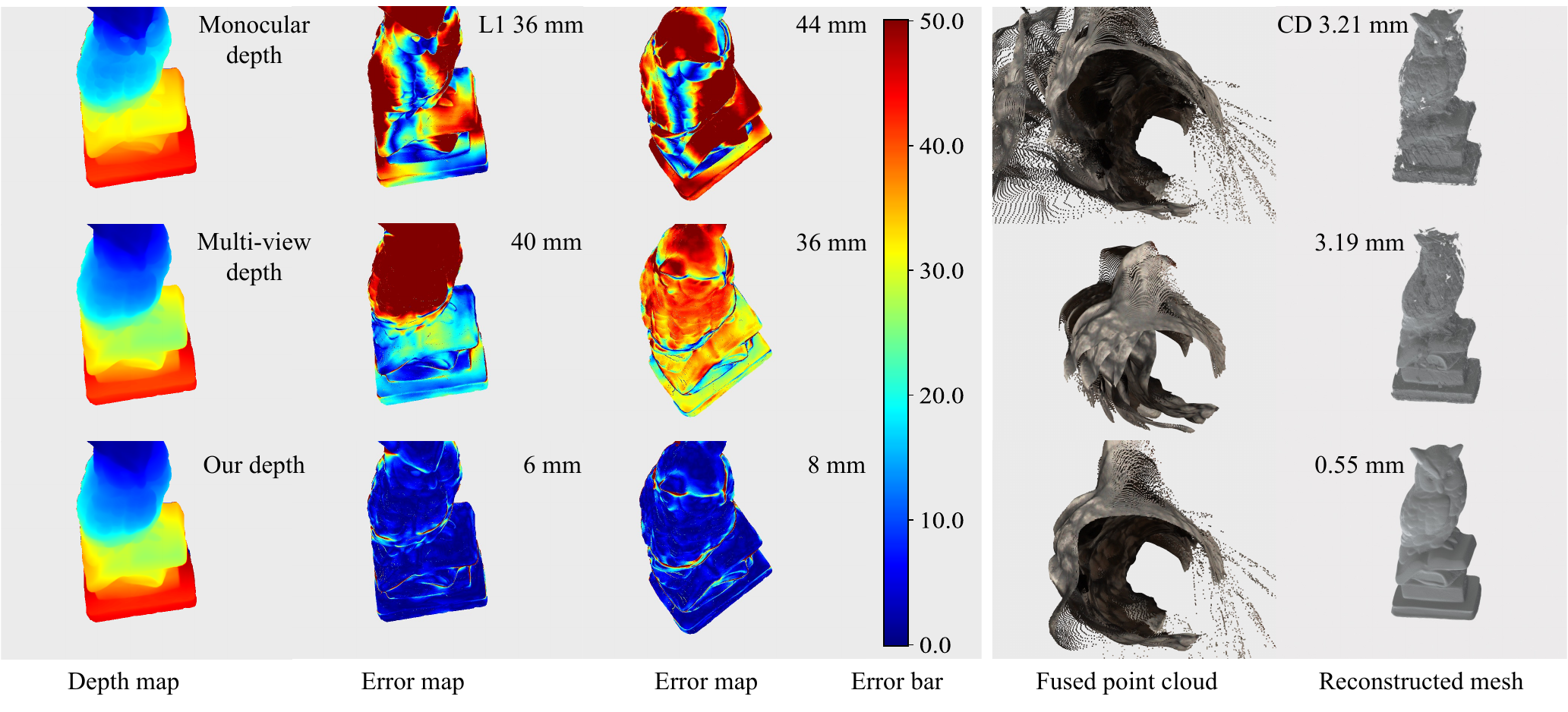}
  \caption{Depth accuracy drives surface quality. The per-view–corrected monocular and multi-view depth maps look plausible, yet L1 distance comparison against GOF’s \cite{yu2024gaussian} geo-consistent depth renderings reveals $\approx$40 mm residual error. Fusing these depths into point clouds exposes severe cross-view mis-alignment that propagates into rough, hole-ridden meshes with high Chamfer distance error. Our SwiftNDC cuts the depth error to less than 10 mm, producing a point cloud that aligns tightly across views and a final surface that is visually smooth and metrically accurate.
}
  \label{fig:depth_pcd}
\end{figure*}

\subsection{Initial depth estimation and correction}\label{sec:perview}

Let $\mathcal{I}=\{I_i\}_{i=1}^{N}$ be the input images, with intrinsics $\mathbf{K}_i$ and extrinsics $(\mathbf{R}_i,\mathbf{t}_i)$ from COLMAP, and let $\mathcal{P}=\{\mathbf{x}_k\}_{k=1}^{M}$ denote the sparse COLMAP point cloud.  Our objective in this subsection is to derive, for each view $i$, an initially corrected depth map $\widetilde{D}_i$ that is aligned to $\mathcal{P}$ and contains as much geometric detail as possible.  This map seeds the pixel-level refinement of Section~\ref{sec:depth_correction_field}.

We first feed the entire image set to VGGT, which jointly reasons over all views and outputs a globally consistent depth map
\begin{equation}
D_i^{\mathrm{vggt}}\colon\Omega_i\!\to\!\mathbb{R}_{+},
\end{equation}
where $\Omega_i\subset\mathbb{R}^{2}$ is the pixel grid.  Processing every image simultaneously eliminates most of the scale drift found in earlier $K$-view MVS nets, but VGGT tends to oversmooth thin structures and exhibits local bias. To recover lost high-frequency geometry, we independently run the VDA monocular model on each frame, producing $D_i^{\mathrm{mono}}$.  These depths capture geometry details well but are only relative; their per-view scales may differ.  Importantly, VGGT and VDA run in seconds for $N<60$, so combining them incurs negligible overhead.

Next, each 3D point $\mathbf{x}_k$ visible in view $i$ projects to

\begin{equation}
\mathbf{p}_{ik}= \pi(\mathbf{K}_i,\mathbf{R}_i,\mathbf{t}_i;\mathbf{x}_k),\quad
d_{ik}^{\mathrm{col}}=\bigl(\mathbf{R}_i\mathbf{x}_k+\mathbf{t}_i\bigr)_{z},
\end{equation}
providing an accurate depth anchor.  At the same pixel we sample
$d_{ik}^{\mathrm{vggt}}=D_i^{\mathrm{vggt}}(\mathbf{p}_{ik})$
and
$d_{ik}^{\mathrm{mono}}=D_i^{\mathrm{mono}}(\mathbf{p}_{ik})$,
yielding the triplet set

\begin{equation}
\mathcal{T}_i=\!\bigl\{(d_{ik}^{\mathrm{vggt}},\,d_{ik}^{\mathrm{mono}},\,d_{ik}^{\mathrm{col}})\bigr\}_{k=1}^{K_i},
\end{equation}
with $K_i\approx10^{3}\!-\!10^{4}$. We align both depth maps to the sparse point cloud by solving two independent ordinary-least-squares problems:

\begin{equation}
\min_{s,b}\sum_{k=1}^{K_i}\bigl(s\,d_{ik}^{t}+b-d_{ik}^{\mathrm{col}}\bigr)^{2},
\quad t\in\{\mathrm{vggt},\mathrm{mono}\},
\end{equation}
which admit analytic solutions in $\mathcal{O}(K_i)$ time.  The affine-corrected depths are

\begin{align}
\widetilde{D}_i^{\mathrm{vggt}}(\mathbf{p}) &= s_i^{\mathrm{v}}D_i^{\mathrm{vggt}}(\mathbf{p})+b_i^{\mathrm{v}}, \\
\widetilde{D}_i^{\mathrm{mono}}(\mathbf{p}) &= s_i^{\mathrm{m}}D_i^{\mathrm{mono}}(\mathbf{p})+b_i^{\mathrm{m}}.
\end{align}

These maps weave together VGGT’s global consistency and monocular fine detail but still contain residual local error (see Fig.~\ref{fig:depth_pcd}). We next introduce a sparse-guided, pixel-level correction field that removes those errors and unlocks high-fidelity geometry.

\subsection{Neural depth correction}\label{sec:depth_correction_field}

The per-view affine alignment rectifies gross scale drift, yet centimetre-level, spatially varying errors remain inside each depth map. When these residuals are integrated along viewing rays in a TSDF volume they amplify into surface ripples and pinholes that defeat the goal of a reconstruction (see Fig.~\ref{fig:depth_pcd}). We therefore introduce a neural depth correction field that predicts a pixel-wise affine residual from nothing more than the sparse, accurate COLMAP anchors.

For every sparse point $\mathbf x_k$ observed in view $i$ we collect two depth samples and three view descriptors, forming

\begin{equation}
\mathbf z_{ik} =
\bigl(d^{\mathrm v}_{ik},\, d^{\mathrm m}_{ik},\, u_{ik},\, v_{ik},\, \iota_i\bigr)\in\mathbb R^{5},
\end{equation}
where $d^{\mathrm v}_{ik}=\widetilde D^{\mathrm{vggt}}_i(\mathbf p_{ik})$ and
$d^{\mathrm m}_{ik}=\widetilde D^{\mathrm{mono}}_i(\mathbf p_{ik})$ are the affine-corrected VGGT and monocular depths, $(u_{ik},v_{ik})\in[-1,1]^2$ are normalized image-plane coordinates, and $\iota_i=i/(N-1)$ normalizes the view index.  An MLP $f_{\boldsymbol\theta}$ with six hidden layers of width 64 maps this five-vector with  positional encoding $\varphi$ to four coefficients,

\begin{equation}
\bigl(\alpha_{ik}^{\mathrm v},\beta_{ik}^{\mathrm v},
       \alpha_{ik}^{\mathrm m},\beta_{ik}^{\mathrm m}\bigr)
    = f_{\boldsymbol\theta}(\varphi(\mathbf z_{ik})),
\end{equation}
and delivers the corrected depths
\begin{equation}
\hat d^{\mathrm v}_{ik}=e^{\alpha_{ik}^{\mathrm v}}\,
                       d^{\mathrm v}_{ik}+\beta_{ik}^{\mathrm v},
\qquad
\hat d^{\mathrm m}_{ik}=e^{\alpha_{ik}^{\mathrm m}}\,
                       d^{\mathrm m}_{ik}+\beta_{ik}^{\mathrm m}.
\end{equation}
Exponentiating the learned scales guarantees positivity without extra constraints.  We train the network for view $i$ by minimizing the sparse L1 reprojection loss
\begin{equation}\label{eq:loss}
\mathcal L(\boldsymbol\theta)= 
\sum_{k\in\mathcal T_i}
\left(
\lvert\hat d^{\mathrm v}_{ik}-d^{\mathrm{col}}_{ik}\rvert
+
\lvert\hat d^{\mathrm m}_{ik}-d^{\mathrm{col}}_{ik}\rvert
\right),
\end{equation}
where $d^{\mathrm{col}}_{ik}$ is the COLMAP depth.

Directly minimizing the above loss per view requires about 5000 gradient steps per image, or several minutes for a 60-view scene.  We therefore amortize training across views by introducing a global-then-local schedule. We first learn a scene-level field by solving

\begin{equation*}
\min_{\boldsymbol\theta}\;
\mathcal L_{\mathrm{global}}(\boldsymbol\theta)=
\sum_{i=1}^{N}\sum_{k\in\mathcal T_i}
\!\!
\left(
\lvert\hat d^{\mathrm v}_{ik}-d^{\mathrm{col}}_{ik}\rvert
+
\lvert\hat d^{\mathrm m}_{ik}-d^{\mathrm{col}}_{ik}\rvert
\right),
\end{equation*}
using AdamW \cite{loshchilovdecoupled} with an initial learning rate of $1\times10^{-3}$ and the cosine annealing warm restart scheduler \cite{loshchilov2022sgdr} (restart period $T_0=1000$, total steps $T=5000$).  This global optimization captures systematic biases shared by all views. The converged weights $\boldsymbol\theta^\star$ serve as a warm start for a per-view refinement using Eq.~\ref{eq:loss} again with AdamW, a restart period $T_0=250$, and only 500 steps, typically finishing in less than one second. As shown in Fig.~\ref{fig:depth_pcd}, this per-view correction significantly reduces depth error and, in turn, produces accurate mesh that cannot be achieved using feed-forward depths alone.

\subsection{Reliable dense geometry initialization}
After obtaining corrected depths $\hat{D}_i$, we construct a dense point cloud by back-projecting each pixel in view $i$. For a pixel $\mathbf{p}_{iu}$, the corresponding 3D point is
\begin{equation}
\mathbf{x}_{iu}=\Pi(\mathbf{K}_i,\mathbf{R}_i,\mathbf{t}_i; \mathbf{p}_{iu}),
\label{eq:backproj}
\end{equation}
where $\Pi$ is the back-projection operation. Although the corrected depth maps are aligned, individual pixels may still contain residual spatial errors. We therefore evaluate each 3D point's reliability using a view-to-view reprojection cycle. We reproject the 3D point into a neighboring view $j$:
\begin{equation}
\mathbf{p}_{ju} = \pi(\mathbf{K}_j,\mathbf{R}_j,\mathbf{t}_j;\mathbf{x}_{iu}),
\end{equation}
At this 2D location $\mathbf{p}_{ju}$, we sample the corrected depth $\hat{D}_j(\mathbf{p}_{ju})$, back-project it in view $j$ to obtain $\mathbf{x}_{ju}$, and then reproject this 3D point back into view $i$ to get the reprojected pixel location $\hat{\mathbf{p}}_{iu}$. We define the reprojection error in pixel space as
\begin{equation}
e_{i\rightarrow j}(\mathbf{p}_{iu})=\left\| \hat{\mathbf{p}}_{iu} - \mathbf{p}_{iu} \right\|_2.
\label{eq:reproj_error}
\end{equation}

\begin{figure}[t]
  \centering
  \includegraphics[width=\linewidth]{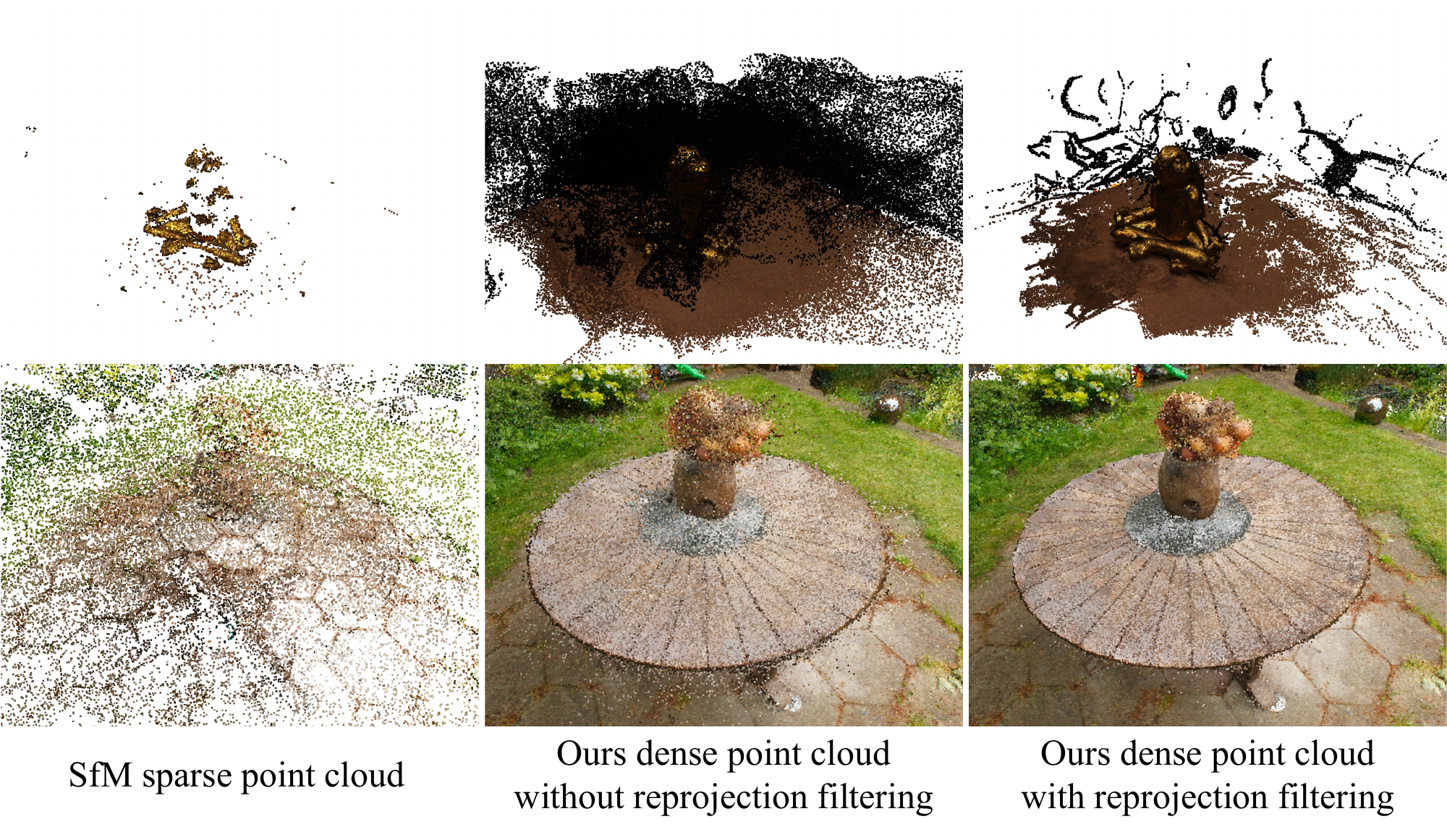}
  \caption{Effect of removing unreliable points using reprojection-error filtering.}
  \label{fig:reprojection}
\end{figure}

A point is considered reliable if its average reprojection error across a small set of overlapping views is below a threshold. We set this threshold as 1 pixel to ensure reliability. The filtered set of points forms a clean and geometrically consistent dense point cloud, as illustrated in Fig.~\ref{fig:reprojection}. We further apply a simple downsampling method to obtain a uniformly distributed subset, which serves as the dense geometry initialization for 3DGS. This initialization yields a high-quality geometric scaffold that can significantly accelerate 3DGS convergence for mesh reconstruction and improve rendering quality for novel view synthesis.

\section{Experiments}\label{sec:exp}

\subsection{Experiment settings}\label{sec:settings}

\begin{table*}[t]
\centering
\caption{Surface reconstruction accuracy on the DTU dataset~\cite{jensen2014large}, measured by Chamfer distance $\downarrow$.
Ours-Depth directly reconstructs meshes from the corrected depth maps, while Ours-3DGS further applies a lightweight 3DGS refinement (1k iterations, initialized from our dense point samples) to enhance surface fidelity. Results are highlighted as \colorbox{red}{1st}, \colorbox{orange}{2nd}, and \colorbox{yellow}{3rd}.}
\label{tab:dtu_results}
\begin{adjustbox}{width=\textwidth,center}
\begin{tabular}{lccccccccccccccccr}
\toprule
\textbf{Method} & 24 & 37 & 40 & 55 & 63 & 65 & 69 & 83 & 97 & 105 & 106 & 110 & 114 & 118 & 122 & \textbf{Mean} & \textbf{Time} \\ 
\midrule
NeRF~\cite{mildenhall2020nerf} & 1.90 & 1.60 & 1.85 & 0.58 & 2.28 & 1.27 & 1.47 & 1.67 & 2.05 & 1.07 & 0.88 & 2.53 & 1.06 & 1.15 & 0.96 & 1.49 & $>$12h \\
VolSDF~\cite{yariv2021volume} & 1.14 & 1.26 & 0.81 & 0.49 & 1.25 & \cellcolor{yellow}0.70 & 0.72 & 1.29 & 1.18 & 0.70 & 0.66 & 1.08 & 0.42 & 0.61 & 0.55 & 0.86 & $>$12h \\
NeuS~\cite{wang2021neus} & 1.00 & 1.37 & 0.93 & 0.43 & 1.10 & \cellcolor{orange}0.65 & \cellcolor{orange}0.57 & 1.48 & 1.09 & 0.83 & \cellcolor{orange}0.52 & 1.20 & \cellcolor{yellow}0.35 & \cellcolor{yellow}0.49 & 0.54 & 0.84 & $>$12h \\
\midrule
3DGS~\cite{kerbl20233d} & 2.14 & 1.53 & 2.08 & 1.68 & 3.49 & 2.21 & 1.43 & 2.07 & 2.22 & 1.75 & 1.79 & 2.55 & 1.53 & 1.52 & 1.50 & 1.96 & 20m \\
SuGaR~\cite{guedon2024sugar} & 1.47 & 1.33 & 1.13 & 0.61 & 2.25 & 1.71 & 1.15 & 1.63 & 1.62 & 1.07 & 0.79 & 2.45 & 0.98 & 0.88 & 0.79 & 1.33 & 107m \\
2DGS~\cite{huang20242d} & 0.48 & 0.91 & 0.39 & \cellcolor{yellow}0.39 & 1.01 & 0.83 & 0.81 & 1.36 & 1.27 & 0.76 & 0.70 & 1.40 & 0.40 & 0.76 & 0.52 & 0.80 & \cellcolor{yellow}19m \\
GOF~\cite{yu2024gaussian} & 0.50 & 0.82 & \cellcolor{yellow}0.37 & \cellcolor{orange}0.37 & 1.12 & 0.74 & 0.73 & \cellcolor{yellow}1.18 & 1.29 & \cellcolor{yellow}0.68 & 0.77 & \cellcolor{yellow}0.90 & 0.42 & 0.66 & \cellcolor{yellow}0.49 & \cellcolor{yellow}0.74 & 33m \\
PGSR~\cite{chen2024pgsr} & \cellcolor{red}0.34 & \cellcolor{red}0.58 & \cellcolor{red}0.29 & \cellcolor{red}0.29 & \cellcolor{red}0.78 & \cellcolor{red}0.58 & \cellcolor{red}0.54 & \cellcolor{red}1.01 & \cellcolor{red}0.73 & \cellcolor{red}0.51 & \cellcolor{red}0.49 & \cellcolor{orange}0.69 & \cellcolor{red}0.31 & \cellcolor{red}0.37 & \cellcolor{red}0.38 & \cellcolor{red}0.53 & 30m \\
\midrule
Ours depth & \cellcolor{yellow}0.44 & \cellcolor{orange}0.60 & \cellcolor{orange}0.37 & 0.41 & \cellcolor{yellow}0.91 & 1.01 & 0.61 & 1.80 & \cellcolor{yellow}1.00 & 0.69 & 0.64 & 1.28 & 0.44 & 0.56 & 0.55 & 0.75 & \cellcolor{red}1m \\
Ours 3DGS & \cellcolor{orange}0.39 & \cellcolor{yellow}0.68 & 0.46 & 0.39 & \cellcolor{orange}0.80 & 0.73 & \cellcolor{yellow}0.59 & \cellcolor{orange}1.06 & \cellcolor{orange}0.84 & \cellcolor{orange}0.59 & \cellcolor{yellow}0.58 & \cellcolor{red}0.67 & \cellcolor{orange}0.33 & \cellcolor{orange}0.42 & \cellcolor{orange}0.40 & \cellcolor{orange}0.59 & \cellcolor{orange}3m \\
\bottomrule
\end{tabular}
\end{adjustbox}
\end{table*}

\begin{table}[h!]
    \centering
    \caption{Quantitative results of F1 score $\uparrow$ for surface reconstruction on Tanks and Temples dataset \cite{knapitsch2017tanks}. Our method attains comparable reconstruction quality to existing approaches while requiring the shortest reconstruction time.}
    \label{tab:tnt}
    \begin{adjustbox}{width=\linewidth,center}
        \begin{tabular}{lccccc}
            \hline
             & Neurlangelo & 2DGS~\cite{huang20242d} & GOF~\cite{yu2024gaussian} & PGSR~\cite{chen2024pgsr} & Ours\\
            \hline
            Barn            & \cellcolor{red}0.70 & 0.36 & 0.51 & \cellcolor{orange}0.66  & \cellcolor{yellow}0.62 \\
            Caterpillar     & \cellcolor{yellow}0.36  & 0.23 & \cellcolor{orange}0.41 & \cellcolor{orange}0.41 & \cellcolor{red}0.42 \\
            Courthouse      & \cellcolor{red}0.28  & 0.13 & \cellcolor{red}0.28 & \cellcolor{yellow}0.21 & \cellcolor{orange}0.25 \\
            Ignatius        & \cellcolor{red}0.89  & 0.44 & \cellcolor{yellow}0.68 & \cellcolor{orange}0.80 & \cellcolor{orange}0.80 \\
            Meetingroom     & \cellcolor{red}0.32  & 0.16 & 0.28 & \cellcolor{yellow}0.29 & \cellcolor{orange}0.31 \\
            Truck           & \cellcolor{yellow}0.48 & 0.26 & \cellcolor{orange}0.58 & \cellcolor{red}0.60 & \cellcolor{red}0.60 \\
            \hline
            Mean            & \cellcolor{red}0.50  & \cellcolor{yellow}0.30 & \cellcolor{orange}0.46 & \cellcolor{red}0.50 & \cellcolor{red}0.50\\
            Time            & $>$128h & \cellcolor{orange}35m & 2h & \cellcolor{yellow}45m & \cellcolor{red}26m \\
            \hline
        \end{tabular}
    \end{adjustbox}
\end{table}

Our neural depth correction field is implemented in Tiny CUDA Neural Networks \cite{muller2021tcnn} for high arithmetic throughput. Corrected depths can be directly fused into a TSDF to extract surface mesh by Marching Cubes \cite{zhou2018open3d} or used to produce reliable dense geometry initialization. For mesh reconstruction, we adopt the DTU surface-reconstruction benchmark \cite{jensen2014large} and the Tanks and Temples (TnT) dataset \cite{knapitsch2017tanks}. Thanks to our reliable dense geometry initialization, our method requires only 1k 3DGS optimization steps to achieve state-of-the-art mesh quality on the DTU benchmark. For large-scale scenes in TnT, we use 15k steps, approximately 50\% of the iterations typically required for full 3DGS optimization. We compare against two families of state-of-the-art baselines: neural implicit methods (NeRF \cite{mildenhall2020nerf}, VolSDF \cite{yariv2021volume}, and NeuS \cite{wang2021neus}) and 3DGS methods (original 3DGS \cite{kerbl20233d}, SuGaR \cite{guedon2024sugar}, 2DGS \cite{huang20242d}, GOF \cite{yu2024gaussian} and PGSR \cite{chen2024pgsr}). All training timing results are measured on a single NVIDIA L40S with 48 GB VRAM. Training times are averaged across all the scenes. 

For view synthesis, we evaluate on three widely used multi-view benchmarks: MipNeRF 360 \cite{barron2022mip}, Tanks and Temples \cite{knapitsch2017tanks}, and Deep Blending~\cite{hedman2018deep}. We adopt the standard train–test splits and report PSNR, SSIM, and LPIPS to assess rendering quality. Baseline comparison includes NeRF-based methods \cite{muller2022instant,barron2021mip,barron2022mip} and recent 3DGS variants \cite{kerbl20233d,tancik2023nerfstudio} using their recommended hyperparameters.

\subsection{Mesh reconstruction}
Table~\ref{tab:dtu_results} reports CD surface error on the DTU scans together with averaged per-scene optimization time. The neural‐implicit baselines deliver competitive geometry (mean errors 0.84–1.49 mm) but require more than twelve hours of per-scene optimization, an order of magnitude longer than most explicit approaches. Gaussian splatting slashes runtime to the minutes range while typically improving accuracy. For instance, GOF achieves the mean error 0.74 mm in 33 minutes. In comparison, our method without any 3DGS optimization already achieves a mean CD of 0.75~mm in only 60 seconds, making it comparable to the best explicit methods while being 20–30 times faster. Viewed on the speed-quality front, this marks the first point that lies inside the sub-minute regime without conceding accuracy to minutes-long pipelines. With an additional lightweight 3DGS refinement, our method further reduces error, achieving results comparable to PGSR (0.59 vs. 0.53 mm) while still running an order of magnitude faster.

\begin{figure*}[t]
  \centering
  \includegraphics[width=\textwidth]{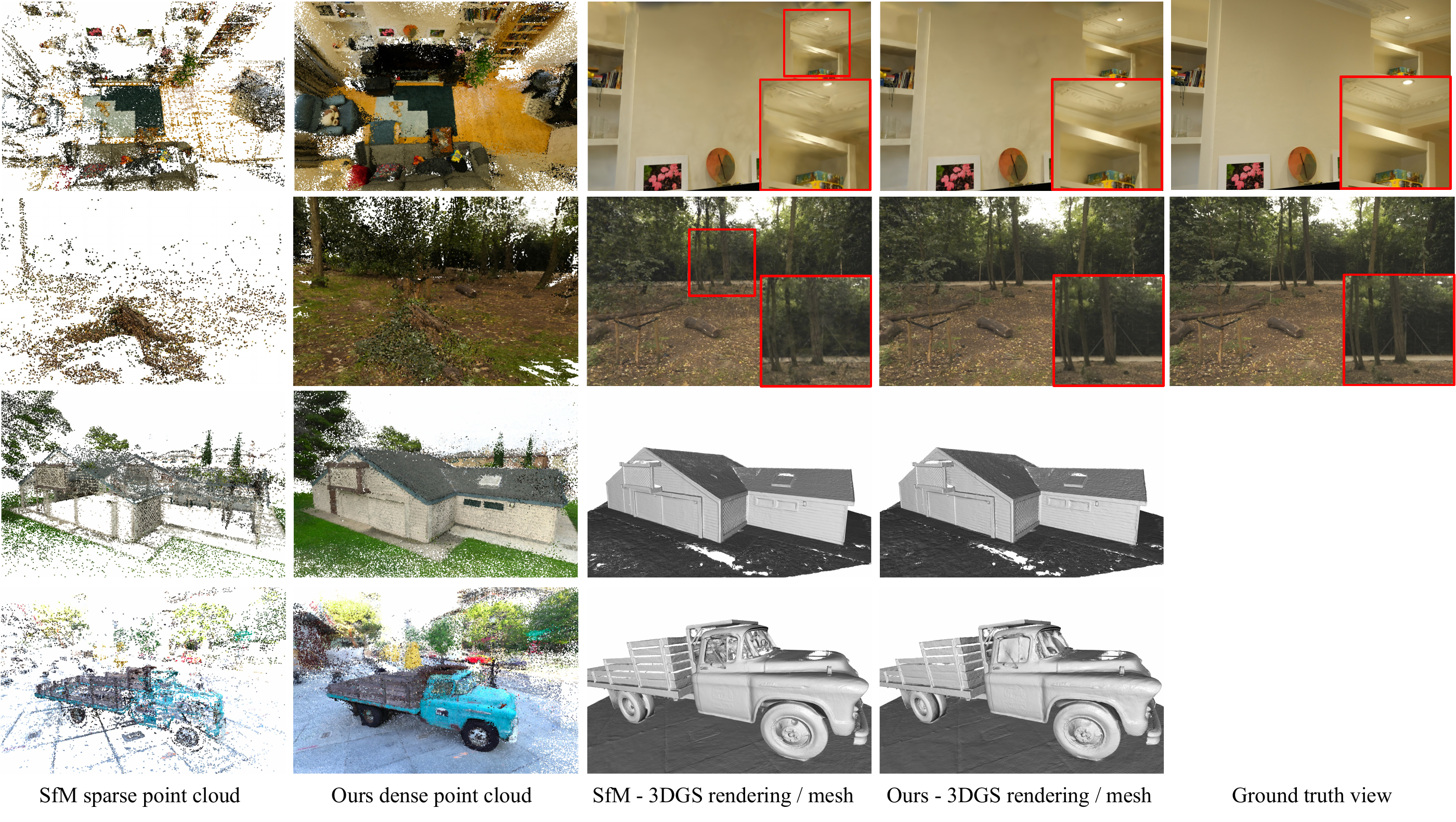}
  \caption{Qualitative results for view synthesis and mesh reconstruction. Our dense initialization yields more complete geometry, especially in sparsely observed regions, enabling higher-quality rendering or reduced  mesh reconstruction time with similar quality.}
  \label{fig:view_mesh_comparison}
\end{figure*}

We present surface reconstruction results on the TnT benchmark, evaluated using the F1 score in Table~\ref{tab:tnt}. Across all six scenes, our method consistently matches or exceeds the performance of recent 3DGS approaches while achieving a mean F1 of 0.50, on par with the two strongest Neurlangelo and PGSR. Crucially, this accuracy is obtained with the shortest reconstruction time, requiring only 26 minutes, substantially faster than PGSR (45 minutes), GOF (2 hours), and orders of magnitude faster than Neurlangelo (over 128 hours). Qualitative results in Fig.~\ref{fig:view_mesh_comparison} further illustrate that our method produces dense point clouds that are both geometrically accurate and well distributed, clearly improving over the sparse SfM input. This dense and reliable point geometry provides a strong initialization for subsequent 3DGS mesh refinement, enabling our final meshes to closely match the quality of PGSR despite much shorter optimization schedules.

\begin{table*}[h]
\centering
\caption{View synthesis quality comparison. Our method delivers consistent improvements across three datasets and all quality metrics.}
\label{tab:view_syn_results}
\begin{adjustbox}{width=0.85\linewidth,center}
\begin{tabular}{lccccccccc}
\toprule
& \multicolumn{3}{c}{MipNeRF 360 \cite{barron2022mip}} & \multicolumn{3}{c}{Tank \& Temples \cite{knapitsch2017tanks}} & \multicolumn{3}{c}{Deep Blending\cite{hedman2018deep}} \\
            & PSNR$\uparrow$      & SSIM$\uparrow$     & LPIPS$\downarrow$    & PSNR$\uparrow$       & SSIM$\uparrow$      & LPIPS$\downarrow$      & PSNR$\uparrow$       & SSIM$\uparrow$     & LPIPS$\downarrow$     \\
\midrule
Instant-NGP \cite{muller2022instant} & 26.75     & 0.75     & 0.30     & 21.92      & 0.75      & 0.31       &            & -        &           \\
MipNeRF \cite{barron2021mip}    & 27.60     & 0.81     & 0.25     &            & -         &            & 21.54      & 0.78     & 0.37      \\
MipNeRF 360 \cite{barron2022mip}  & \cellcolor{orange}29.23     & \cellcolor{yellow}0.84     & \cellcolor{orange}0.21     & 22.22      & 0.76      & 0.26       &            & -        &           \\
\midrule
3DGS (SfM sparse) ~\cite{kerbl20233d}        & 28.69     & \cellcolor{orange}0.87     & \cellcolor{yellow}0.22     & \cellcolor{yellow}23.14      & \cellcolor{yellow}0.84      & \cellcolor{yellow}0.21       &            & -        &           \\
Splatfacto (SfM sparse) \cite{tancik2023nerfstudio}  & \cellcolor{yellow}29.05     & \cellcolor{orange}0.87     & \cellcolor{orange}0.21     & \cellcolor{orange}23.51      & \cellcolor{orange}0.85      & \cellcolor{orange}0.19       & \cellcolor{orange}29.45      & \cellcolor{orange}0.90     & \cellcolor{orange}0.31      \\
Ours (Sampled dense, splatfacto)       & \cellcolor{red}29.33     & \cellcolor{red}0.88     & \cellcolor{red}0.19     & \cellcolor{red}23.74      & \cellcolor{red}0.86      & \cellcolor{red}0.17       & \cellcolor{red}29.74      & \cellcolor{red}0.91     & \cellcolor{red}0.30 \\
\bottomrule
\end{tabular}
\end{adjustbox}
\end{table*}

\subsection{View synthesis}

We summarize the novel view synthesis performance in Table~\ref{tab:view_syn_results}. Across all three datasets and all evaluation metrics, our method consistently outperforms both NeRF-based baselines and recent 3DGS pipelines built on sparse SfM points. When integrated into Splatfacto \cite{tancik2023nerfstudio}, our sampled dense initialization yields the best scores on every benchmark, improving PSNR and substantially reducing perceptual error (LPIPS). The qualitative results in Fig.~\ref{fig:view_mesh_comparison} reveal why: compared to the sparse SfM input, our dense point clouds provide significantly more complete geometric coverage, including regions where the original SfM reconstruction contains few or no points. This reliable and uniformly distributed initialization enables the subsequent 3DGS optimization to better model scene structures that are weakly observed in the training images and would otherwise remain under-densified. As a result, our method boosts rendering quality especially in challenging areas with occlusions, grazing angles, or limited viewpoint diversity. These results also suggest strong compatibility with existing 3DGS variants. Our dense sampling acts as a drop-in enhancement that can be paired with different splatting pipelines to systematically improve view synthesis fidelity.

\subsection{Ablation study}\label{sec:ablation}

Table~\ref{tab:ablation} dissects how each design choice affects accuracy and runtime.  The full neural depth correction attains the mean error 0.75 mm in one minute, establishing the reference point. \textbf{Per-view correction}. Per-view corrected monocular variant by the same SfM points reaches 3.15 mm mean CD and the multi-view variant 2.50 mm, significantly worse than the full model despite taking only a few seconds.  These results confirm that per-view correction cannot resolve cross-view mis-alignment. Pixel-level correction is therefore essential for high-fidelity reconstruction. \textbf{Dual-depth synergy}. Correcting monocular depth (Monocular depth head) rather than multi-view depth increases error to 0.91 mm, while using either modality in isolation (Monocular only, Multi-view only) pushes the error to approximately 1.0 mm. Combining the two cues therefore yields a 25 \% accuracy gain without significantly hurting speed.

\textbf{Two-stage training schedule}.
If the neural depth correction field is trained separately for each image (W/o two-stage training), the error remains competitive (0.81 mm) yet optimization balloons to 12 minutes, twelve times slower than the proposed amortized scheme.  The global pass thus captures most of the bias at once, allowing each fine-tune to converge in less than one second.
\textbf{Runtime observations}.
Using a single depth branch trims 5–10 \% off the one-minute budget (0.89–0.94 min), but the accuracy drop outweighs the marginal savings.  Conversely, the two-stage schedule slashes computation by an order of magnitude with no loss in fidelity, demonstrating that where we spend optimization budget matters more than how much we spend.

\textbf{Reliable dense geometry initialization}. While a lightweight 3DGS refinement can further reduce the mean surface error, it is effective only when built upon our reliable dense geometry initialization. Without this module, the error becomes even larger than that of directly fusing depth maps, underscoring the crucial role of our outlier filtering strategy in enabling fast and high-quality mesh reconstruction. To summarize, every modeling cue, including global VGGT depth, local monocular detail, pixel-level correction and reliable dense geometry initialization, contributes measurably, and their benefits are maximally realized within our two-stage optimization framework.

\begin{table}[t]
  \centering
  \caption{Ablation study on the DTU dataset.}
  \label{tab:ablation}
  \begin{tabular}{l r r}
    \toprule
    Model &  Mean CD $\downarrow$ & Time (m) \\
    \midrule
    Full neural depth correction  & 0.75 & 1 \\
    \midrule
    Per-view cor. mo. depth & 3.15 & 0.18 \\
    Per-view cor. mv. depth & 2.50 & 0.13 \\
    \midrule
    Monocular depth head    & 0.91 & 1 \\
    Monocular depth only    & 1.00 & 0.94   \\
    Multi-view depth only   & 0.98 & 0.89   \\
    W/o two-stage training  & 0.81 & 12     \\
    \midrule
    3DGS opt w reliable geo. & 0.59 & 3 \\
    3DGS opt w/o reliable geo. & 0.82 & 3 \\
    \bottomrule
  \end{tabular}
\end{table}

\begin{table}
  \centering
  \caption{Reconstruction time breakdown (seconds) on the DTU scenes with 49 images.}
  \label{tab:runtime}
  \begin{tabular}{l r}
    \toprule
    Stage & Time (s) \\
    \midrule
    VGGT depth inference              & 3.6 \\
    Monocular depth inference          & 6.6 \\
    Initial per-view correction        & 7.0 \\
    Global scale refinement            & 7.0 \\
    Per-pixel correction         & 30.9 \\
    3DGS optimization-1k iterations         & 110 \\
    TSDF fusion + Marching Cubes       & 1.6 \\
    \midrule
    Total                     & 167 \\
    \bottomrule
  \end{tabular}
\end{table}

\subsection{Reconstruction time breakdown}
Table~\ref{tab:runtime} shows that a complete \SwiftNDC reconstruction finishes in just under a minute. More than half of this time (about 31 s) is spent on the per-pixel refinement network, which adjusts depth values with pixel-level precision. The remaining steps are comparatively lightweight: depth inference from VGGT and the monocular model together take roughly 10 seconds, while the two scale-correction passes consume another 14 seconds. TSDF integration followed by Marching Cubes completes in less than two seconds, indicating that surface extraction is not a time bottleneck once accurate depths are available.

\subsection{Limitations}
While SwiftNDC produces reliable geometry and improves both the efficiency and the quality of 3DGS, it still depends on accurate SfM poses and on the quality of the initial monocular or MVS depth maps. Although SwiftNDC significantly reduces the number of 3DGS iterations for mesh reconstruction, it introduces additional preprocessing steps, including depth correction and reprojection-based filtering, that may become noticeable for extremely large image collections (e.g., thousands of views). Finally, SwiftNDC uses the corrected depths only to initialize 3DGS, and the two stages are not jointly optimized. An end-to-end formulation that couples depth correction with radiance-field optimization could potentially yield even higher fidelity and represents an interesting direction for future work.
\section{Conclusion}
This work demonstrates that reliable dense geometry initialization greatly enhances fast mesh reconstruction and high-quality view synthesis. By transforming sparse SfM points into a dense, accurate, and well-distributed geometric prior, our method provides a strong foundation for downstream 3DGS without the need for long optimization schedules. The proposed approach integrates global multi-view depth, fine-grained monocular detail, and rigorous outlier filtering into a unified dense initialization that is both robust and highly efficient to compute.

This initialization alone yields competitive surfaces on the DTU dataset. With further 3DGS refinement, it matches the mesh quality of state-of-the-art methods at a fraction of their runtime on large-scale outdoor scenes. For view synthesis, the dense geometry fills regions poorly covered by SfM, enabling more consistent and robust rendering across datasets. Overall, our approach demonstrates that fast, scalable 3D reconstruction is best achieved by coupling efficient dense geometry with targeted Gaussian refinement, offering a simple and compatible drop-in improvement for existing 3DGS methods.

\section*{Acknowledgments}
This research was supported by the Australian Research Council Discovery Project (Grant Number DP240103334).

{
    \small
    \bibliographystyle{ieeenat_fullname}
    \bibliography{main}

@String(TOG= {ACM Trans. Graph.})

@String(AAAI = {AAAI})

@String(TOG   = {ACM TOG})

@inproceedings{park2019deepsdf,
  title={{DeepSDF}: Learning continuous signed distance functions for shape representation},
  author={Park, Jeong Joon and Florence, Peter and Straub, Julian and Newcombe, Richard and Lovegrove, Steven},
  booktitle={Proceedings of the IEEE/CVF Conference on Computer Vision and Pattern Recognition},
  pages={165--174},
  year={2019}
}

@inproceedings{wang2021neus,
  title={{NeuS}: learning neural implicit surfaces by volume rendering for multi-view reconstruction},
  author={Wang, Peng and Liu, Lingjie and Liu, Yuan and Theobalt, Christian and Komura, Taku and Wang, Wenping},
  booktitle={Proceedings of the 35th International Conference on Neural Information Processing Systems},
  pages={27171--27183},
  year={2021}
}

@article{yariv2021volume,
  title={Volume rendering of neural implicit surfaces},
  author={Yariv, Lior and Gu, Jiatao and Kasten, Yoni and Lipman, Yaron},
  journal={Advances in Neural Information Processing Systems},
  volume={34},
  pages={4805--4815},
  year={2021}
}

@inproceedings{li2023neuralangelo,
  title={Neuralangelo: High-fidelity neural surface reconstruction},
  author={Li, Zhaoshuo and M{\"u}ller, Thomas and Evans, Alex and Taylor, Russell H and Unberath, Mathias and Liu, Ming-Yu and Lin, Chen-Hsuan},
  booktitle={Proceedings of the IEEE/CVF Conference on Computer Vision and Pattern Recognition},
  pages={8456--8465},
  year={2023}
}

@inproceedings{oechsle2021unisurf,
  title={Unisurf: Unifying neural implicit surfaces and radiance fields for multi-view reconstruction},
  author={Oechsle, Michael and Peng, Songyou and Geiger, Andreas},
  booktitle={Proceedings of the IEEE/CVF international conference on computer vision},
  pages={5589--5599},
  year={2021}
}

@inproceedings{mildenhall2020nerf,
  title={{NeRF}: representing scenes as neural radiance fields for view synthesis},
  author={Mildenhall, Ben and Srinivasan, Pratul P and Tancik, Matthew and Barron, Jonathan T and Ramamoorthi, Ravi and Ng, Ren},
  booktitle={European Conference on Computer Vision},
  pages={405--421},
  year={2020},
  organization={Springer}
}

@article{kerbl20233d,
  title={{3D} {Gaussian} splatting for real-time radiance field rendering},
  author={Kerbl, Bernhard and Kopanas, Georgios and Leimkuehler, Thomas and Drettakis, George},
  journal={ACM Transactions on Graphics (TOG)},
  volume={42},
  number={4},
  pages={1--14},
  year={2023},
  publisher={ACM New York, NY, USA}
}

@inproceedings{guedon2024sugar,
  title={{SuGaR}: Surface-aligned gaussian splatting for efficient {3D} mesh reconstruction and high-quality mesh rendering},
  author={Gu{\'e}don, Antoine and Lepetit, Vincent},
  booktitle={Proceedings of the IEEE/CVF Conference on Computer Vision and Pattern Recognition},
  pages={5354--5363},
  year={2024}
}

@inproceedings{huang20242d,
  title={{2D} {Gaussian} splatting for geometrically accurate radiance fields},
  author={Huang, Binbin and Yu, Zehao and Chen, Anpei and Geiger, Andreas and Gao, Shenghua},
  booktitle={SIGGRAPH'24: ACM SIGGRAPH 2024 Conference Papers},
  volume={32},
  year={2024},
  organization={ACM}
}

@inproceedings{jensen2014large,
  title={Large scale multi-view stereopsis evaluation},
  author={Jensen, Rasmus and Dahl, Anders and Vogiatzis, George and Tola, Engin and Aan{\ae}s, Henrik},
  booktitle={Proceedings of the IEEE/CVF Conference on Computer Vision and Pattern Recognition},
  pages={406--413},
  year={2014}
}

@article{yu2024gaussian,
  title={Gaussian opacity fields: Efficient adaptive surface reconstruction in unbounded scenes},
  author={Yu, Zehao and Sattler, Torsten and Geiger, Andreas},
  journal={ACM Transactions on Graphics (TOG)},
  volume={43},
  number={6},
  pages={1--13},
  year={2024},
}

@inproceedings{zhangneural,
  title={Neural Signed Distance Function Inference through Splatting {3D} Gaussians Pulled on Zero-Level Set},
  author={Zhang, Wenyuan and Liu, Yu-Shen and Han, Zhizhong},
  booktitle={The Thirty-eighth Annual Conference on Neural Information Processing Systems},
  year={2024}
}

@inproceedings{yao2018mvsnet,
  title={{MVSNet}: Depth inference for unstructured multi-view stereo},
  author={Yao, Yao and Luo, Zixin and Li, Shiwei and Fang, Tian and Quan, Long},
  booktitle={Proceedings of the European Conference on Computer Vision},
  pages={767--783},
  year={2018}
}

@inproceedings{yao2019recurrent,
  title={Recurrent {MVSNet} for high-resolution multi-view stereo depth inference},
  author={Yao, Yao and Luo, Zixin and Li, Shiwei and Shen, Tianwei and Fang, Tian and Quan, Long},
  booktitle={Proceedings of the IEEE/CVF Conference on Computer Vision and Pattern Recognition},
  pages={5525--5534},
  year={2019}
}

@inproceedings{gu2020cascade,
  title={Cascade cost volume for high-resolution multi-view stereo and stereo matching},
  author={Gu, Xiaodong and Fan, Zhiwen and Zhu, Siyu and Dai, Zuozhuo and Tan, Feitong and Tan, Ping},
  booktitle={Proceedings of the IEEE/CVF Conference on Computer Cision and Pattern Recognition},
  pages={2495--2504},
  year={2020}
}

@inproceedings{ding2022transmvsnet,
  title={Trans{MVSNet}: Global context-aware multi-view stereo network with transformers},
  author={Ding, Yikang and Yuan, Wentao and Zhu, Qingtian and Zhang, Haotian and Liu, Xiangyue and Wang, Yuanjiang and Liu, Xiao},
  booktitle={Proceedings of the IEEE/CVF Conference on Computer Vision and Pattern Recognition},
  pages={8585--8594},
  year={2022}
}

@article{caomvsformer,
  title={{MVSFormer}: Multi-View Stereo by Learning Robust Image Features and Temperature-based Depth},
  author={Cao, Chenjie and Ren, Xinlin and Fu, Yanwei},
  journal={Transactions on Machine Learning Research},
  year={2022}
}

@inproceedings{caomvsformerpp,
  title={{MVSFormer}++: Revealing the Devil in Transformer's Details for Multi-View Stereo},
  author={Cao, Chenjie and Ren, Xinlin and Fu, Yanwei},
  booktitle={The Twelfth International Conference on Learning Representations},
  year={2024}
}

@inproceedings{wolf2024gs2mesh,
  title={{GS2Mesh}: Surface reconstruction from gaussian splatting via novel stereo views},
  author={Wolf, Yaniv and Bracha, Amit and Kimmel, Ron},
  booktitle={European Conference on Computer Vision},
  pages={207--224},
  year={2024},
  organization={Springer}
}

@inproceedings{schonberger2016structure,
  title={Structure-from-motion revisited},
  author={Schonberger, Johannes L and Frahm, Jan-Michael},
  booktitle={Proceedings of the IEEE Conference on Computer Vision and Pattern Recognition},
  pages={4104--4113},
  year={2016}
}

@inproceedings{schonberger2016pixelwise,
  title={Pixelwise view selection for unstructured multi-view stereo},
  author={Sch{\"o}nberger, Johannes L and Zheng, Enliang and Frahm, Jan-Michael and Pollefeys, Marc},
  booktitle={European Conference on Computer Vision},
  pages={501--518},
  year={2016},
  organization={Springer}
}

@inproceedings{wang2025vggt,
  title={{VGGT}: Visual geometry grounded transformer},
  author={Wang, Jianyuan and Chen, Minghao and Karaev, Nikita and Vedaldi, Andrea and Rupprecht, Christian and Novotny, David},
  booktitle={Proceedings of the Computer Vision and Pattern Recognition Conference},
  pages={5294--5306},
  year={2025}
}

@inproceedings{li2018megadepth,
  title={{MegaDepth}: Learning single-view depth prediction from internet photos},
  author={Li, Zhengqi and Snavely, Noah},
  booktitle={Proceedings of the IEEE Conference on Computer Vision and Pattern Recognition},
  pages={2041--2050},
  year={2018}
}

@inproceedings{bhat2021adabins,
  title={{AdaBins}: Depth estimation using adaptive bins},
  author={Bhat, Shariq Farooq and Alhashim, Ibraheem and Wonka, Peter},
  booktitle={Proceedings of the IEEE Conference on Computer Vision and Pattern Recognition},
  pages={4009--4018},
  year={2021}
}

@inproceedings{yan2023desnet,
  title={{DesNet}: Decomposed scale-consistent network for unsupervised depth completion},
  author={Yan, Zhiqiang and Wang, Kun and Li, Xiang and Zhang, Zhenyu and Li, Jun and Yang, Jian},
  booktitle={Proceedings of the AAAI Conference on Artificial Intelligence},
  volume={37},
  number={3},
  pages={3109--3117},
  year={2023}
}

@inproceedings{izquierdo2023sfm,
  title={{SfM-TTR}: Using structure from motion for test-time refinement of single-view depth networks},
  author={Izquierdo, Sergio and Civera, Javier},
  booktitle={Proceedings of the IEEE/CVF Conference on Computer Vision and Pattern Recognition},
  pages={21466--21476},
  year={2023}
}

@article{fink2024refinement,
  title={Refinement of Monocular Depth Maps via Multi-View Differentiable Rendering},
  author={Fink, Laura and Franke, Linus and Keinert, Joachim and Stamminger, Marc},
  journal={arXiv preprint arXiv:2410.03861},
  year={2024}
}

@inproceedings{li2024radarcam,
  title={{RadarCam-depth}: Radar-camera fusion for depth estimation with learned metric scale},
  author={Li, Han and Ma, Yukai and Gu, Yaqing and Hu, Kewei and Liu, Yong and Zuo, Xingxing},
  booktitle={2024 IEEE International Conference on Robotics and Automation (ICRA)},
  pages={10665--10672},
  year={2024},
  organization={IEEE}
}

@inproceedings{ganj2025hybriddepth,
  title={{HybridDepth}: Robust Metric Depth Fusion by Leveraging Depth from Focus and Single-Image Priors},
  author={Ganj, Ashkan and Su, Hang and Guo, Tian},
  booktitle={2025 IEEE/CVF Winter Conference on Applications of Computer Vision (WACV)},
  pages={973--982},
  year={2025},
  organization={IEEE}
}

@software{muller2021tcnn,
    author = {M\"uller, Thomas},
    license = {BSD-3-Clause},
    month = {4},
    title = {{tiny-cuda-nn}},
    url = {https://github.com/NVlabs/tiny-cuda-nn},
    version = {1.7},
    year = {2021}
}

@inproceedings{chen2025video,
  title={Video depth anything: Consistent depth estimation for super-long videos},
  author={Chen, Sili and Guo, Hengkai and Zhu, Shengnan and Zhang, Feihu and Huang, Zilong and Feng, Jiashi and Kang, Bingyi},
  booktitle={Proceedings of the Computer Vision and Pattern Recognition Conference},
  pages={22831--22840},
  year={2025}
}

@inproceedings{curless1996volumetric,
  title={A volumetric method for building complex models from range images},
  author={Curless, Brian and Levoy, Marc},
  booktitle={Proceedings of the 23rd Annual Conference on Computer Graphics and Interactive Techniques},
  pages={303--312},
  year={1996}
}

@article{zhou2018open3d,
  title={{Open3D}: A modern library for 3D data processing},
  author={Zhou, Qian-Yi and Park, Jaesik and Koltun, Vladlen},
  journal={arXiv preprint arXiv:1801.09847},
  year={2018}
}

@article{farshian2023deep,
  title={Deep-learning-based {3D} surface reconstruction—a survey},
  author={Farshian, Anis and G{\"o}tz, Markus and Cavallaro, Gabriele and Debus, Charlotte and Nie{\ss}ner, Matthias and Benediktsson, J{\'o}n Atli and Streit, Achim},
  journal={Proceedings of the IEEE},
  volume={111},
  number={11},
  pages={1464--1501},
  year={2023},
  publisher={IEEE}
}

@inproceedings{loshchilovdecoupled,
  title={Decoupled Weight Decay Regularization},
  author={Loshchilov, Ilya and Hutter, Frank},
  booktitle={International Conference on Learning Representations},
  year={2017}
}

@inproceedings{loshchilov2022sgdr,
  title={SGDR: Stochastic Gradient Descent with Warm Restarts},
  author={Loshchilov, Ilya and Hutter, Frank},
  booktitle={International Conference on Learning Representations},
  year={2022}
}

@article{knapitsch2017tanks,
  title={Tanks and temples: Benchmarking large-scale scene reconstruction},
  author={Knapitsch, Arno and Park, Jaesik and Zhou, Qian-Yi and Koltun, Vladlen},
  journal={ACM Transactions on Graphics},
  volume={36},
  number={4},
  pages={1--13},
  year={2017},
  publisher={ACM New York, NY, USA}
}

@article{chen2024pgsr,
  title={{PGSR}: Planar-based gaussian splatting for efficient and high-fidelity surface reconstruction},
  author={Chen, Danpeng and Li, Hai and Ye, Weicai and Wang, Yifan and Xie, Weijian and Zhai, Shangjin and Wang, Nan and Liu, Haomin and Bao, Hujun and Zhang, Guofeng},
  journal={IEEE Transactions on Visualization and Computer Graphics},
  year={2024},
  publisher={IEEE}
}

@inproceedings{barron2022mip,
  title={{Mip-NeRF} 360: Unbounded anti-aliased neural radiance fields},
  author={Barron, Jonathan T and Mildenhall, Ben and Verbin, Dor and Srinivasan, Pratul P and Hedman, Peter},
  booktitle={Proceedings of the IEEE/CVF conference on computer vision and pattern recognition},
  pages={5470--5479},
  year={2022}
}

@article{hedman2018deep,
  title={Deep blending for free-viewpoint image-based rendering},
  author={Hedman, Peter and Philip, Julien and Price, True and Frahm, Jan-Michael and Drettakis, George and Brostow, Gabriel},
  journal={ACM Transactions on Graphics (ToG)},
  volume={37},
  number={6},
  pages={1--15},
  year={2018},
  publisher={ACM New York, NY, USA}
}

@article{muller2022instant,
  title={Instant neural graphics primitives with a multiresolution hash encoding},
  author={M{\"u}ller, Thomas and Evans, Alex and Schied, Christoph and Keller, Alexander},
  journal={ACM transactions on graphics (TOG)},
  volume={41},
  number={4},
  pages={1--15},
  year={2022},
  publisher={ACM New York, NY, USA}
}

@inproceedings{tancik2023nerfstudio,
  title={Nerfstudio: A modular framework for neural radiance field development},
  author={Tancik, Matthew and Weber, Ethan and Ng, Evonne and Li, Ruilong and Yi, Brent and Wang, Terrance and Kristoffersen, Alexander and Austin, Jake and Salahi, Kamyar and Ahuja, Abhik and others},
  booktitle={ACM SIGGRAPH 2023 conference proceedings},
  pages={1--12},
  year={2023}
}

@article{kotovenko2025edgs,
  title={{EDGS}: Eliminating Densification for Efficient Convergence of 3DGS},
  author={Kotovenko, Dmytro and Grebenkova, Olga and Ommer, Bj{\"o}rn},
  journal={arXiv preprint arXiv:2504.13204},
  year={2025}
}

@inproceedings{barron2021mip,
  title={{Mip-NeRF}: A multiscale representation for anti-aliasing neural radiance fields},
  author={Barron, Jonathan T and Mildenhall, Ben and Tancik, Matthew and Hedman, Peter and Martin-Brualla, Ricardo and Srinivasan, Pratul P},
  booktitle={Proceedings of the IEEE/CVF international conference on computer vision},
  pages={5855--5864},
  year={2021}
}
}


\end{document}